\documentclass[11pt,english]{article}
\usepackage[T1]{fontenc}
\usepackage[latin9]{inputenc}
\usepackage{geometry}
\usepackage{amsmath}
\usepackage{bbm}
\usepackage{babel}
\usepackage{graphicx}
\usepackage{subcaption}
\usepackage{xcolor}
\usepackage{booktabs}
\usepackage{dsfont}
\usepackage{multirow}
\usepackage{floatrow}
\newfloatcommand{capbtabbox}{table}[][\FBwidth]
\usepackage{diagbox}
\definecolor{amethyst}{rgb}{0.7, 0.4, 0.8}
\definecolor{asparagus}{rgb}{0.53, 0.66, 0.42}
\definecolor{charcoal}{rgb}{0.21, 0.27, 0.31}

\usepackage{hyperref}
\hypersetup{
    colorlinks=true,
    linkcolor=blue,
    filecolor=magenta,      
    urlcolor=cyan,
}
\usepackage[title]{appendix}

\usepackage{listings}
\usepackage{color}
 
\definecolor{codegreen}{rgb}{0,0.6,0}
\definecolor{codegray}{rgb}{0.5,0.5,0.5}
\definecolor{codepurple}{rgb}{0.58,0,0.82}
\definecolor{backcolour}{rgb}{0.95,0.95,0.92}
 
\lstdefinestyle{mystyle}{
    backgroundcolor=\color{backcolour},   
    commentstyle=\color{codegreen},
    keywordstyle=\color{magenta},
    numberstyle=\tiny\color{codegray},
    stringstyle=\color{codepurple},
    basicstyle=\footnotesize,
    breakatwhitespace=false,         
    breaklines=true,                 
    captionpos=b,                    
    keepspaces=true,                 
    numbers=left,                    
    numbersep=5pt,                  
    showspaces=false,                
    showstringspaces=false,
    showtabs=false,                  
    tabsize=2
}
\usepackage[round]{natbib}
\begin{document}
\title{The Evolution of Popularity and Images of Characters in \\ Marvel Cinematic Universe Fanfictions}
\author{Fan Bu}
\date{}
\maketitle

\begin{abstract}
This analysis proposes a new topic model to study the yearly trends in Marvel Cinematic Universe fanfictions on three levels: character popularity, character images/topics, and vocabulary pattern of topics. It is found that character appearances in fanfictions have become more diverse over the years thanks to constant introduction of new characters in feature films, and in the case of Captain America, multi-dimensional character development is well-received by the fanfiction world.
\end{abstract}

\section{Introduction}
\textbf{Fanfictions} are fictions written by fans as creative responses to books, comics, TV shows, movies, etc. out of appreciation and passion for the storytelling and/or the main characters. Authors often write about the characters or a character's personality aspects they find most inspiring or relatable, so the contents of fanfictions can reflect the popularity and fans' perception of the characters involved.

The \textbf{Marvel Cinematic Universe (MCU)} is an American media franchise and shared universe that is centered on a series of superhero films, produced by Marvel Studios and based on characters in comic books published by Marvel Comics.\footnote{Paraphrased from the ``Marvel Cinematic Universe'' Wikipedia entry.} It has been one of the most commercially successful multimedia franchises in the world and has inspired a large volume of fanfiction writing since 2008.

The profitability and marketability of a franchise like MCU rely heavily on its fanbase, so it is of importance to study the fans - their favorite characters, their perception of these characters, how the perception has evolved through time, etc., and fanfictions may shed some light onto these problems of interest.

This analysis tries to answer these two questions: First, which characters have been most popular in the MCU fanfiction world and have their popularity levels changed over time? Second, what have been the major themes/topics/images depicted about the characters in MCU fanfictions, and how have those themes evolved over time?

\section{Data Collection and Pre-processing}
Textual data of 900 MCU fanfictions, consisting of the top 100 fanfictions (ranked by number of hits) of each year from 2010 to 2018, are collected from \url{archiveofourown.org}\footnote{It is a nonprofit, open source fanfiction repository created in late 2008, hosting 3757000 works and 1448000 users to date.} using a web scraper\footnote{Core codes by Jingyi Li. Github: \url{https://github.com/radiolarian/AO3Scraper}}. For each fanfiction, both the metadata (title, author, publishing date, word count, characters, additional tags, etc.) and the main body of the first chapter are scraped. This dataset contains 6,051,422 words in total.

Natural language pre-processing steps are carried out for the main body text of each fanfiction: digits, punctuations, special symbols, multiple white spaces, non-English words, and English stopwords are removed; all words are stemmed into their vocabulary roots; the main body of each document is split into a list of words.

\section{Exploratory Data Analysis}
\begin{figure}[H]
    \centering
    \begin{subfigure}[t]{0.5\textwidth}
        \centering
        \includegraphics[height=1.3in]{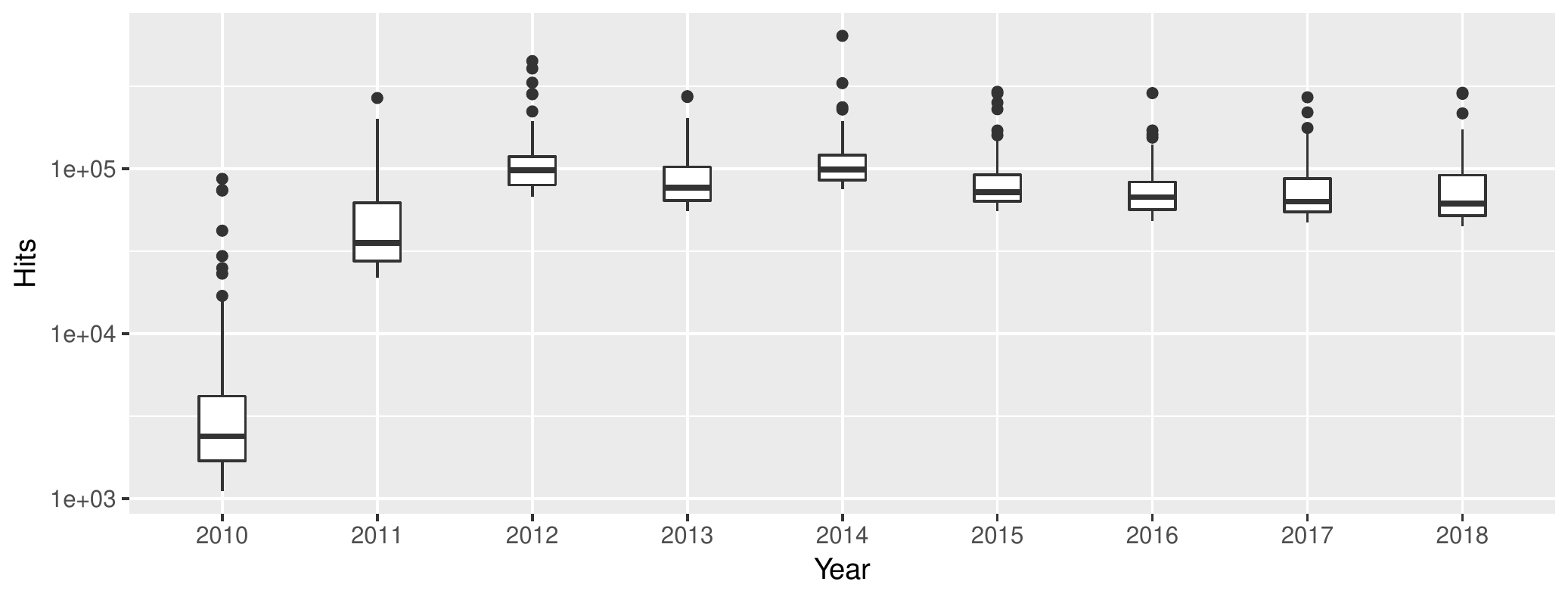}
        \caption{Hits of top 100 fanfictions from 2010 to 2018.}
        \label{fig:hit}
    \end{subfigure}%
    ~ 
    \begin{subfigure}[t]{0.5\textwidth}
        \centering
        \includegraphics[height=1.3in]{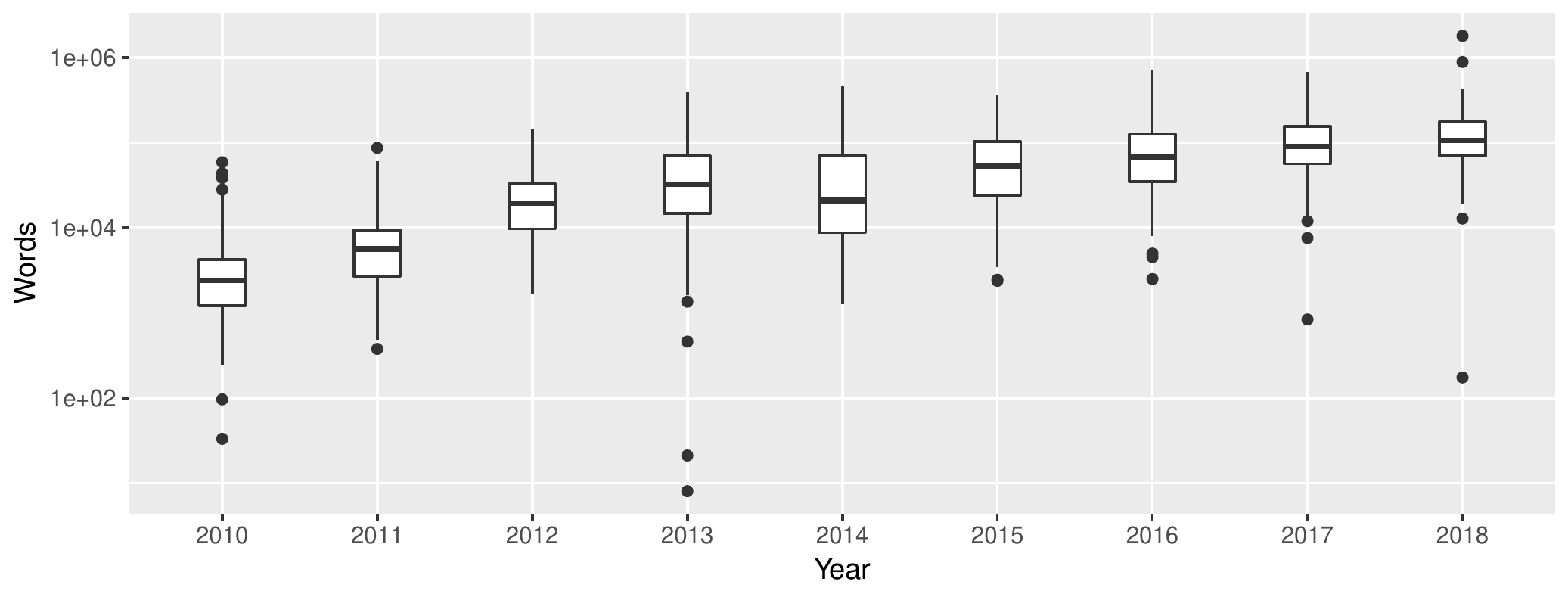}
        \caption{Word counts of top 100 fanfictions from 2010 to 2018.}
        \label{fig:word}
    \end{subfigure}
    \caption{MCU fanfiction reader volume and writing volume over the years.}
\end{figure}

MCU fanfictions demonstrate a high level of engagement from both the readers and the writers. Figure \ref{fig:hit} summarizes the volumes of readers of each year's top 100 fanfictions. From 2012 up to now, an average top 100 fanfiction has approximately 100,000 visits on its page. Figure \ref{fig:word} shows the increasing trend of word counts of top 100 fanfictions over the years. For instance, an average top 100 fanfiction published this year runs about 100,000 words long - therefore, to reduce computational complexity, the collected  dataset only includes the first chapter for each fanfiction.
\vspace*{-0.15in}
Figure 2 visualizes the appearance frequencies in fanfictions for 13 MCU main characters\footnote{They are: Black Widow, Captain America, Winter Soldier, Iron Man, Black Panther, Spiderman, Thor, Loki, Starlord, Dr. Strange, Antman, Hulk, and Deadpool.} throughout the years. For each character, the appearance frequency is obtained by counting the number of fanfictions involving him/her. The top chart shows the appearance frequencies for all 9 years. Cumulatively, Iron Man and Captain America have been the two biggest powerhouses in MCU fanfictions. However, the landscape has changed a lot from 2010 to 2018. In 2010, almost all the works were dedicated to Iron Man, with the other characters barely mentioned, but in 2018 fanfictions are much more diverse in terms of central characters.
\begin{figure}[H]
\centering
\includegraphics[height=7cm]{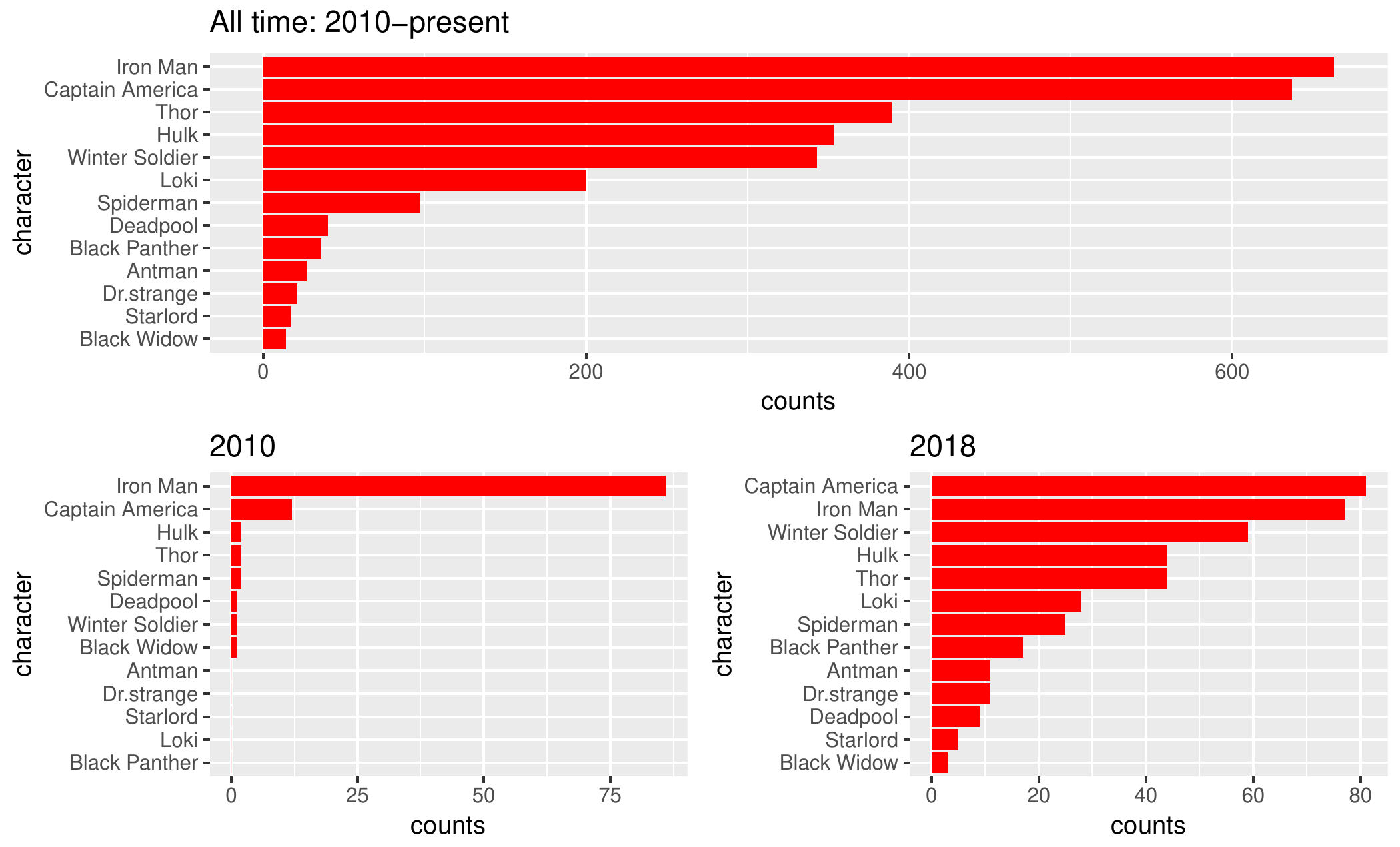}
\caption{Number of fanfictions dedicated to each of 13 MCU main characters. Top: 2010 to present; bottom left: year 2010; bottom right: year 2018.}
\end{figure}
\vspace*{-0.15in}
Although intuitively, the themes/topics associated with a character can be investigated through frequently used words in fanfictions involving them, such a ``raw count'' approach proves to be rather uninformative. Figure 3 presents the word clouds of fanfictions involving Black Widow (Natasha Romanoff) and Hulk (Bruce Banner), with basic stopwords removed. Clearly, most frequent terms are often character names and non-specific common words. Thus, as a further step of processing, those terms are also manually removed. Also, a more sophisticated model seems necessary.
\begin{figure}[H]
    \centering
    \begin{subfigure}[b]{0.5\textwidth}
        \centering
        \includegraphics[height=1.4in]{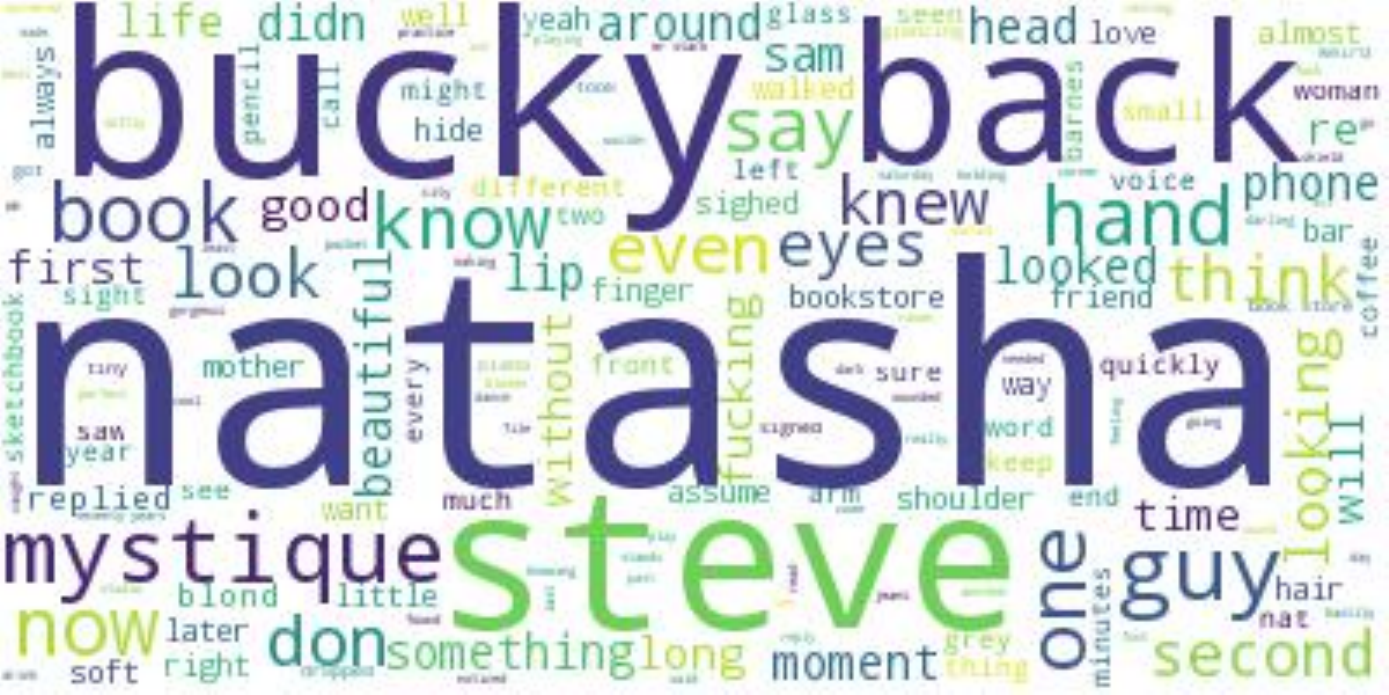}
        \caption{Most frequent words in Black Widow fanfictions.}
    \end{subfigure}%
    ~ 
    \begin{subfigure}[b]{0.5\textwidth}
        \centering
        \includegraphics[height=1.4in]{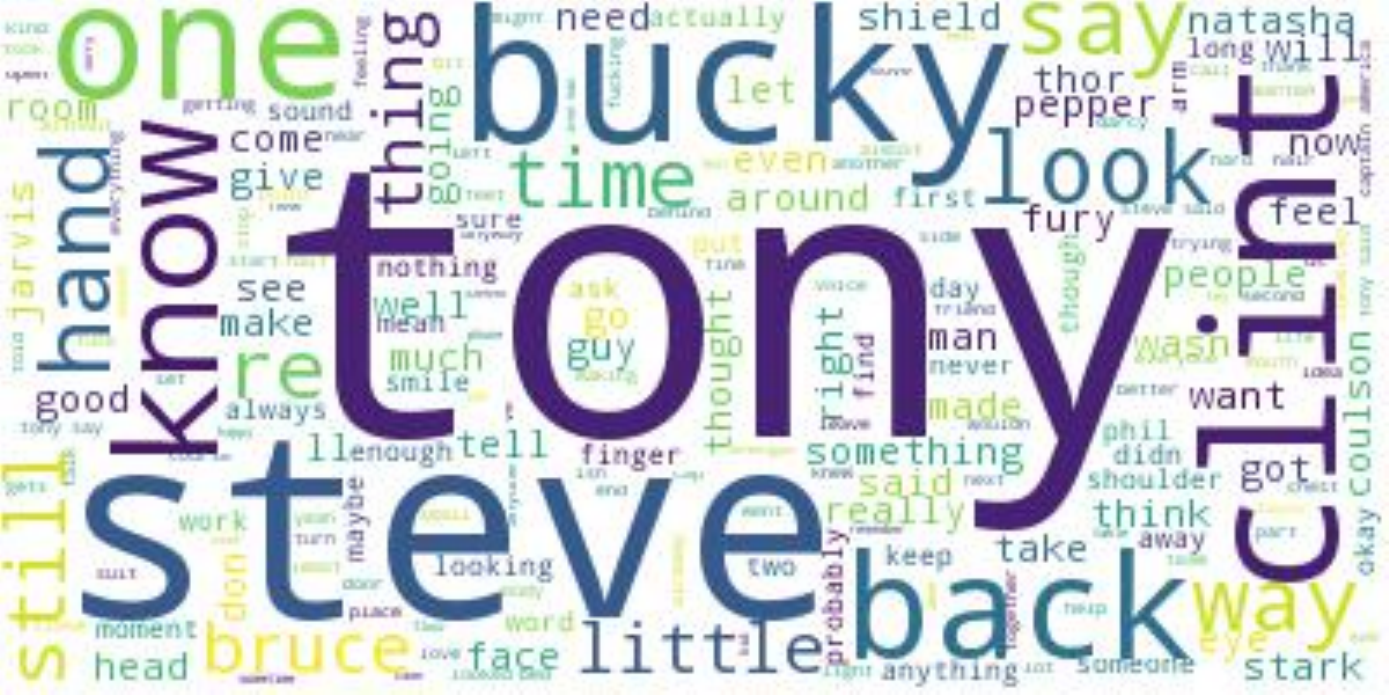}
        \caption{Most frequent words in Hulk fanfictions.}
    \end{subfigure}
    \caption{Word clouds of fanfictions involving Black Widow (Natasha Romanoff) and Hulk (Bruce Banner).}
\end{figure}
\vspace*{-0.3in}

\section{Modeling: Dynamic Labeled Topic Model}
This analysis proposes a new model, the Dynamic Labeled Topic Model, to study the problems of interest. This model can be seen as an extension of the combination of the Dynamic Topic Model \citep{blei2006dynamic} and the Author-Topic Model \citep{rosen2004author}. 
\subsection{Model Specification}
Suppose a document can be represented by a pair of vectors, $(\mathbf{w}_d,\mathbf{l}_d)$, where $\mathbf{w}_d$ is a word vector of length $N_d$ and $\mathbf{l}_d$ is a ``label'' vector of length $L_d$. Here, the labels can be the co-authors of an article, the tags (hashtags) associated with a social media post, or, in the fanfiction setting, the main characters of a fanfiction. Let a collection of documents $\mathcal{D}=\{\mathcal{D}^{(t)}\}$ span over time slots $t = 1,\ldots,T$, in which, at each time slot $t$, $\mathcal{D}^{(t)} = \{(\mathbf{w}_{d^{(t)}},\mathbf{l}_{d^{(t)}})\}_{{d^{(t)}}=1}^{D_t}$ is a collection of $D_t$ documents.

The model is based on the following intuitions/assumptions: First, each textual document is centered around its labels (represented by mixtures of topics) and each label has the same amount of influence on the wording choice; Second, the popularity/frequency of a label may change over time, but within a certain time slot it remains the same; Third, for each label, its associated topic mixture distribution also changes over time, but remains the same within a given time slot; Finally, each topic specifies a particular distribution of words over a total vocabulary, which may change over time but not by much.

The generative process of $\mathcal{D}$ is then: For time slots $t=1:T$,
\vspace*{-0.1in}
\begin{enumerate}
\item Draw time-specific word distribution $\beta_z^{(t)}|\beta_z^{(t-1)} \sim \mathcal{N}(\beta_z^{(t-1)},\sigma^2I)$ for each topic $z$;
\item Draw time-specific ``average'' topic distribution $\alpha^{(t)}|\alpha^{(t-1)} \sim \mathcal{N}(\alpha^{(t-1)},\delta^2I)$
\item Draw label-specific topic distribution $\theta_l^{(t)} \sim \mathcal{N}(\alpha^{(t)},a^2I)$ for each label $l$;
\item Draw label probabilities $\psi^{(t)}|\psi^{(t-1)}\sim Dir(\psi^{(t-1)})$ for time slot $t$;
\item For each document ${d^{(t)}}$, draw its labels $\mathbf{l}_{d^{(t)}} \sim Mult(L_{d^{(t)}};\psi^{(t)})$;
\item For each word $w_{{d^{(t)}},i}$ in document ${d^{(t)}}$,
\begin{enumerate}
\item Sample its label $x_{{d^{(t)}},i}$ uniformly from $\mathbf{l}_{d^{(t)}}$;
\item Draw its topic $z_{{d^{(t)}},i} \sim Mult(\pi(\theta_{x_{{d^{(t)}},i}}^{(t)}))$;
\item Draw the word $w_{{d^{(t)}},i} \sim Mult(\pi(\beta_{z_{{d^{(t)}},i}}^{(t)}))$.
\end{enumerate}
\end{enumerate}
\vspace*{-0.2in}
Here $\pi$ maps the multinomial natural parameters to the mean parameters, e.g. $\pi(\beta_z^{(t)})_w = \frac{\beta_{z,w}^{(t)}}{\sum_w \beta_{z,w}^{(t)}}$.

\subsection{Parameter Inference}
Let $V$ be the total vocabulary size and $Z$ be the number of topics. For each document $d$, let $\tilde{\mathbf{w}}_{d}$ be its bag-of-words representation (a $V$-dimensional vector with each component as the counts of a word). Then the full likelihood given all the parameters is
\begin{equation}
\begin{aligned}
\mathcal{L}(\mathcal{D}) &= \prod_t \prod_{d^{(t)}} Mult(\mathbf{l}_{d^{(t)}} | L_{d^{(t)}};\psi^{(t)}) Mult(\tilde{\mathbf{w}}_{d^{(t)}}|N_{d^{(t)}};\phi^{(t)}_{\mathbf{l}_{d^{(t)}}}) \\
&= \prod_t \prod_{d^{(t)}} Mult(\mathbf{l}_{d^{(t)}} | L_{d^{(t)}};\psi^{(t)})  \times \prod_t \prod_{d^{(t)}} Mult(\tilde{\mathbf{w}}_{d^{(t)}}|N_{d^{(t)}};\phi^{(t)}_{\mathbf{l}_{d^{(t)}}})\\
&=\mathcal{L}_1(\mathcal{D}) \times \mathcal{L}_2(\mathcal{D}),
\end{aligned}
\end{equation}
where
\begin{equation}
\phi^{(t)}_{\mathbf{l}_{d^{(t)}}} = \frac{1}{L_{d^{(t)}}}\sum_{l \in \mathbf{l}_{d^{(t)}}} \frac{1}{Z}\sum_{z=1}^Z \pi(\theta_l^{(t)})_z \pi(\beta_z^{(t)}) 
= \frac{1}{ZL_{d^{(t)}}}\sum_{l \in \mathbf{l}_{d^{(t)}}}\sum_{z=1}^Z \phi_{l,z}^{(t)},
\end{equation}
where $\phi_{l,z}^{(t)}:=\pi(\theta_l^{(t)})_z \pi(\beta_z^{(t)})$, the word probability vector of topic $z$ associated with label $l$ in time slot $t$. Eq (1) and (2) suggest that inference of $\{\psi^{(t)}\}$ should rely on $\{\mathbf{l}_{d^{(t)}}\}$, and that inference of $\{\theta_l^{(t)}\}$ and $\{\beta_z^{(t)}\}$ is essentially inference of $\{\phi_{l,z}^{(t)}\}$ which replies on $\{\tilde{\mathbf{w}}_{d^{(t)}}\}$, and thus the inference can be carried out by optimizing on the two partial likelihoods \citep{cox1975partial}, $\mathcal{L}_1(\mathcal{D})$ and $\mathcal{L}_2(\mathcal{D})$. Moreover, since it is assumed that every label $l$ of a given document $d$ is equally influential on its word choice $\tilde{\mathbf{w}}_{d}$, then a $1/{L_d}$ portion of $\tilde{\mathbf{w}}_{d}$ is expected to ``contribute'' to the inference of $\phi_{l,\cdot}^{(\cdot)}$. 

Thus, the parameters can be inferred approximately through a two-stage algorithm:
\begin{enumerate}
\item[Stage 1] For each time slot $t=1:T$, sample $\psi^{(t)}$ from the posterior $p(\psi^{(t)}|\{\mathbf{l}_{d^{(t)}}\}_{d^{(t)}=1}^{D_t},\{L_{d^{(t)}}\}_{d^{(t)}=1}^{D_t},\psi^{(t-1)})$, which is a Dirichlet by conjugacy;
\item[Stage 2] For each label $l=1:L$, infer parameters $\{\theta_l^{(t)}\}$ and $\{\beta_z^{(t)}\}$ deterministically through an approximate variational posterior\footnote{Variational methods usually serve as deterministic alternatives to stochastic simulation when direct posterior inference is intractable. The idea behind variational methods is to optimize groups of free parameters of a proxy distribution over the target parameters such that the distribution is close to the true posterior in KL divergence.} as proposed by \cite{blei2006dynamic}, with the input bag-of-words vector for each document re-weighted by $1/L_{d}$ for each associated label.
\end{enumerate}

\section{Modeling Results on MCU Fanfictions}
\begin{figure}[H]
    \centering
    \begin{subfigure}[t]{0.5\textwidth}
        \centering
        \includegraphics[height=1.9in]{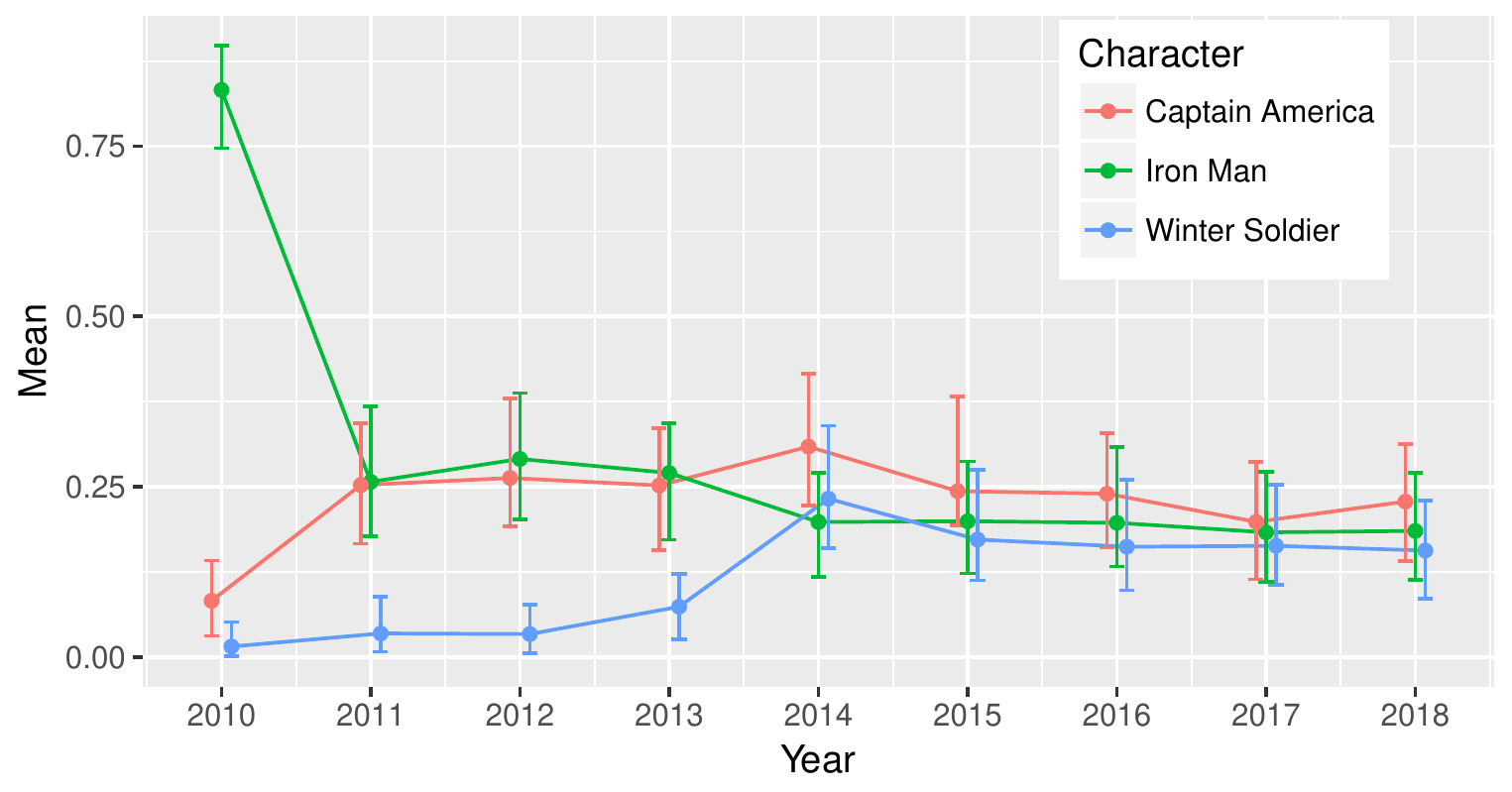}
        \caption{Posterior means and 95\% CIs of the label probabilities for Captain America, Iron Man, and Winter Soldier.}
        \label{fig:Big_Triangle}
    \end{subfigure}%
    ~ 
    \begin{subfigure}[t]{0.5\textwidth}
        \centering
        \includegraphics[height=1.9in]{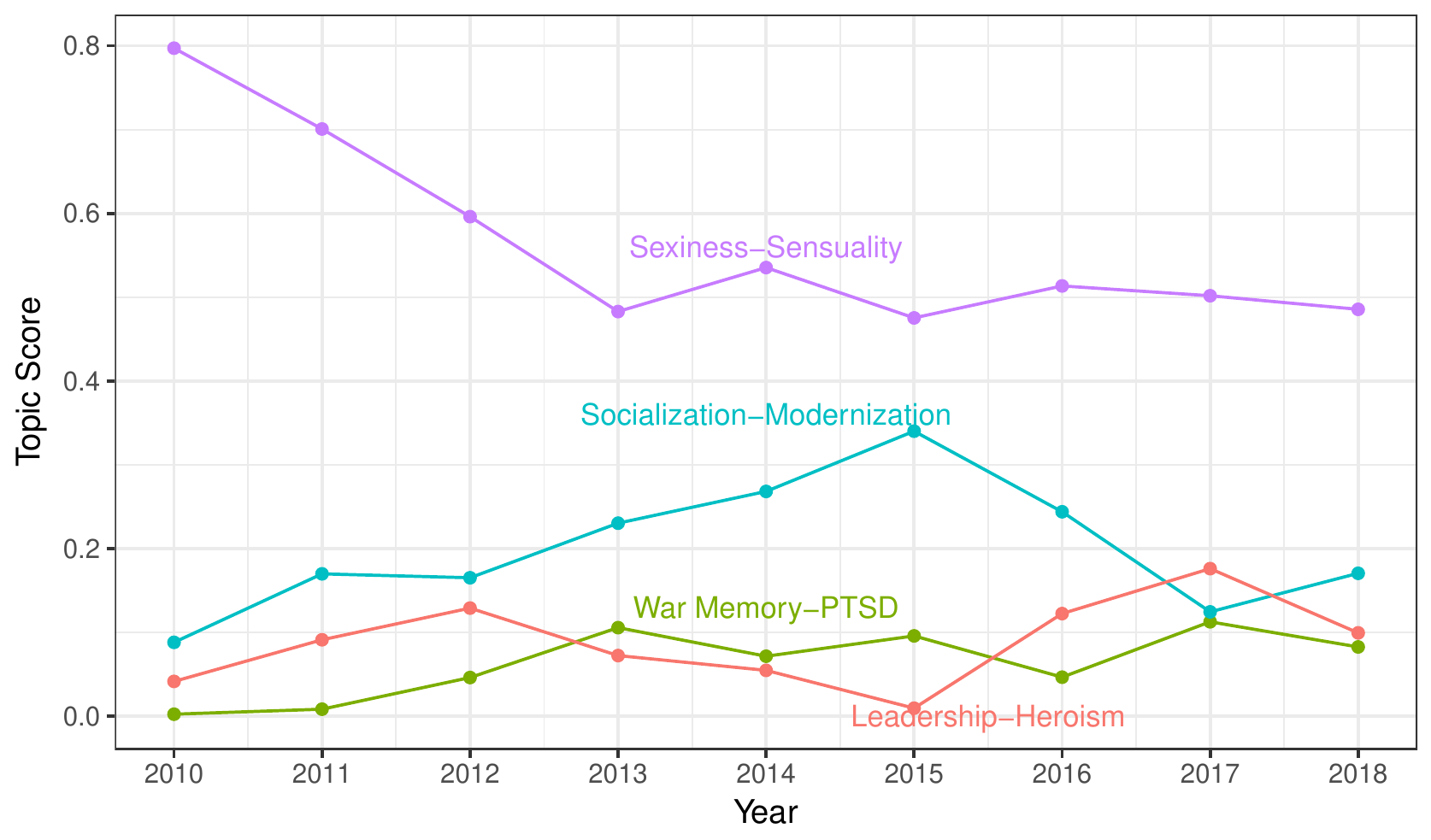}
        \caption{The topic scores of top-4 Captain America topics in fanfictions over the years.}
        \label{fig:Cap_Topic}
    \end{subfigure}
    \caption{Character popularity evolution and Captain America's image evolution.}
\end{figure}

The model proposed in Section 4 is applied to the fanfiction data with vocabulary size $V=27610$, number of ``labels'' (main characters) $L=13$, and number of topics $Z=10$.
\vspace*{-0.2in}
\paragraph{Character Popularity Evolution}
The model learns the label probability for each of the 13 characters in each year, and Figure \ref{fig:Big_Triangle} plots the posterior means and 95\% credible intervals of the label probability components for Captain America, Iron Man, and Winter Soldier. The dominance of Iron Man in MCU fanfictions has been gradually diluted by the arrival of other MCU lead characters (though he is still one of the most written characters in fanfictions); Captain America rose in popularity in 2011 and has remained the most depicted character since 2014; Winter Soldier experienced a major popularity boost in 2014. Such findings are justified by the Marvel feature films timeline: before 2011, MCU had already released 2 Iron Man films, both were major box office successes, but in 2011, the first Captain America film came out so Captain America's popularity started to rise in the fandom, while Winter Soldier was the title character of the second Captain America film in 2014, for which he started receiving more spotlight.

\paragraph{Character Image Evolution} For every main character, his/her associated topics (and their evolution) can be summarized through ``topic scores''. Let the \textbf{topic score} for topic $z$ (associated with label $l$) in time slot $t$ be the estimate of $\pi(\theta_l^{(t)})_z$ and the \textbf{average topic score} for topic $z$ (associated with label $l$) be the average of topic scores over all time slots. Take Captain America as an example. Figure \ref{fig:Cap_Topic} plots the topic scores of the 4 topics associated with Captain America that have highest average topic scores (a table summarizing these 4 topics is given in Appendix A). The \textbf{Sexiness-Sensuality} topic has been an all-time lead, implying that the fandom has always perceived Captain America as a sex symbol, which indeed reflects his physical attractiveness and is also mirrored by certain publicity strategies of Marvel Studios. However, such image has been countered by his kind, friendly personality and engagement in the modern lifestyle, and furthermore, in recent years, fanfictions have been showing more appreciation of his leader/warrior identity and more sympathy for his personal sufferings. Overall, the character of Captain America has become more layered and complicated and the fanfiction world is actively responding to the character development.

\vspace*{-0.3in}

\section{Conclusion and Discussion}
In this analysis, a text corpus of 900 popular MCU fanfictions in the past 9 years was collected, a new textmining model was proposed to study the yearly trends of MCU fanfictions on three levels (popularity of characters, topics associated with characters, and vocabulary pattern of topics), and the model results were interpreted and compared with actual MCU events.

This analysis also provides insights into the fans' creative responses for the benefit of Marvel Studios. Introducing new characters and introducing new depths to existing characters are two essential aspects of developing a cinematic universe. The increased diversity of character depictions in fanfictions reflects success in character introduction, and in the case of Captain America, multi-dimensional character building is well-received and appreciated by fanfiction writers.

It is necessary to point out the limitations of this analysis. First and foremost, despite the large volume of fans engaging in fanfictions, fanfiction writers only account for a small (and skewed, with the majority as female erotica writers) portion of the entire fandom, so the findings should not be generalized without precautions. Moreover, the model specified in Section 4 assumes equal influence from each label on the wording choice, whereas in many scenarios such assumption is invalid (just like there are major characters and minor characters in a story). Future work may consider collecting textual data from other sources (eg. \url{Reddit.com}) as well as non-uniform word-level label assignment in the generative model.

\newpage
\bibliography{ms}

\begin{appendices}

\section{Captain America's Image Summary: Top 4 Topics}
\begin{table}[H]
\centering
\begin{tabular}{ll}
\textbf{Topic}\footnote{Topic names are self-summarized, since topic models are unsupervised and thus do not generate category names.} & \textbf{Frequent Terms}\\
\hline
Sexiness \& Sensuality & **body parts**, **explicit content**\\
Socialization \& Modernization & people, kind, sure, smile, laugh, friend, text, phone \\
Leadership \& Heroism & Avengers, SHIELD, save, New York, armor, suit \\
War Memory \& PTSD & remember, memory, Hydra, kill, hell, blood, war, dead
\end{tabular}
\caption{Top topics and corresponding frequent terms associated with Captain America. Frequent terms are not arranged in a particular order.}
\end{table}

\section{Vocabulary Evolution of Topics}
Since the model allows the word distribution for a specific topic to change over time, we can examine the evolution of frequent terms for a particular topic. Figure 5 shows the top words (ranked by descending word probabilities) of Captain America's ``Socialization \& Modernization'' topic in 2011, 2014, and 2017. It can be observed that the words ``smile'' and ``laugh'' have been used more frequently, probably because he's been perceived as a happier man now than before. Moreover, the word ``phone'' gradually came on top, possibly because he's been getting used to modern technology (or, the fanfiction writers have become more addicted to their own mobile phones).

\vspace*{-0.8in}
\begin{figure}[H]
\centering
\includegraphics[width=11cm]{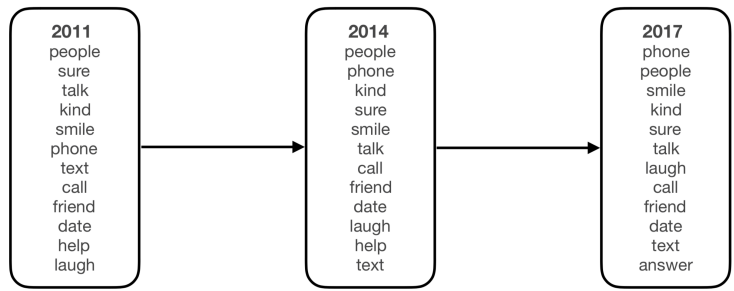}
\vspace*{-0.8in}
\caption{Top words of Captain America's ``Socialization \& Modernization'' topic in 2011, 2014, and 2017.}
\end{figure}

\end{appendices}

\end{document}